\documentclass[conference]{IEEEtran}
\IEEEoverridecommandlockouts
\usepackage{cite}
\usepackage{amsmath,amssymb,amsfonts}
\usepackage{algorithmic}
\usepackage{graphicx}
\usepackage{tabularx}
\usepackage{textcomp}
\usepackage{xcolor}
\usepackage{hyperref}

\newcommand{\gptoss}{\texttt{gpt-oss-20b}~}
\newcommand{\llama}{\texttt{llama-3.3-70b}~}
\newcommand{\qwenlarge}{\texttt{qwen3-30b}~}
\newcommand{\qwensmall}{\texttt{qwen3-4b}~}
\newcommand{\qwenthink}{\texttt{qwen3-4b-thinking}~}

\hypersetup{
    colorlinks=true,
    linkcolor=blue,
    filecolor=magenta,      
    urlcolor=cyan,
    pdftitle={Overleaf Example},
    pdfpagemode=FullScreen,
    }

\def\BibTeX{{\rm B\kern-.05em{\sc i\kern-.025em b}\kern-.08em
    T\kern-.1667em\lower.7ex\hbox{E}\kern-.125emX}}
\begin{document}

\title{Arabic Prompts with English Tools: A Benchmark}

\author{

\IEEEauthorblockN{
Konstantin Kubrak,
Ahmed El-Moselhy,
Ammar Alsulami, 
Remaz Altuwaim,  \\
Hassan Ismail Fawaz and
Faisal Alsaby
}

\IEEEauthorblockA{
GenAI \& Data Analytics Department\\
\textit{General Organization for Social Insurance}\\
Riyadh, Saudi Arabia \\
\{kkubrak,ael-moselhy,aalsulami,raltuwaim,ifawaz,falsaby\}@gosi.gov.sa
}
}

\maketitle

\begin{abstract}
Large Language Models (LLMs) are now integral to numerous industries, increasingly serving as the core reasoning engine for autonomous agents that perform complex tasks through tool-use. 
While the development of Arabic-native LLMs is accelerating, the benchmarks for evaluating their capabilities lag behind, with most existing frameworks focusing on English. 
A critical and overlooked area is tool-calling, where the performance of models prompted in non-English languages like Arabic is poorly understood, especially since these models are often pretrained on predominantly English data. 
This paper addresses this critical gap by introducing the first dedicated benchmark for evaluating the tool-calling and agentic capabilities of LLMs in the Arabic language. 
Our work provides a standardized framework to measure the functional accuracy and robustness of models in Arabic agentic workflows. 
Our findings reveal a huge performance gap: when users interact in Arabic, tool-calling accuracy drops by an average of 5-10\%, regardless of whether the tool descriptions themselves are in Arabic or English. 
By shedding light on these critical challenges, this benchmark aims to foster the development of more reliable and linguistically equitable AI agents for Arabic-speaking users.
\end{abstract}


\section{Introduction}
Large Language Models (LLMs) have precipitated a significant paradigm shift, demonstrating remarkable capabilities that are being leveraged across nearly every industry~\cite{brown2020language}, going from the insurance~\cite{balona2024actuarygpt} and legal~\cite{yang2024beyond} to finance~\cite{patel2024fanal} and social sciences~\cite{sakib2024challenging}.
Beyond text generation, models such as GPT-4 and 5 are increasingly being utilized as the reasoning core for autonomous agents, enabling complex, multi-step agentic workflows~\cite{wang2024survey}, such as leveraging AI Agents (powered by LLMs) in order to perform risk assessment, claims processing, customer engagement and regulatory compliance~\cite{selvadurai2025ai}.

This global trend is mirrored by a growing adoption and development of LLMs tailored for the Arabic language~\cite{mashaabi2024survey}. 
The industry is witnessing a surge in Arabic-centric models and platforms, driven by the need to serve the region's large user base. 
The release of models like Allam~\cite{bari2024allam} and Jais~\cite{sengupta2023jais} signifies a clear move towards high-performance, Arabic-native generative AI.

Despite this proliferation, the evaluation of LLMs remains a significant challenge. 
The majority of prominent benchmarks and leaderboards have been historically English-centric. 
Foundational benchmarks like MMLU~\cite{hendrycks2020measuring} and HELM~\cite{liang2022holistic} primarily test model knowledge and capabilities in English. 
While comprehensive benchmarks for underrepresented languages, including Arabic, have recently begun to appear they primarily focus on traditional NLP tasks such as question answering, summarization, and sentiment analysis~\cite{ghaboura2025arb}.

A key capability of modern LLMs is \textbf{tool calling}, which underpins autonomous agents~\cite{qin2023toolllm} by enabling interaction with external systems, access to real-time information, and execution of complex actions. Evaluation of this capability in multilingual contexts remains limited. Recent work~\cite{rababah2024existing} shows that translation can be exploited for adversarial jailbreaking, suggesting that tool-calling performance may be affected across languages. It is still unclear how prompting models in languages like Arabic impacts their tool-calling accuracy, given that the models and their training data are largely English-based.

We fill this critical gap by introducing a novel benchmark\footnote{The code for this paper is available at: \url{https://github.com/kubrak94/gorilla/}}  designed to evaluate tool-calling and agentic workflow capabilities of LLMs specifically for the Arabic language, based on the famous Berkeley Function Calling Leaderboard (BFCL)~\cite{patil2025bfcl}. 
This work provides a necessary evaluation framework to shed light on the robustness of multilingual models in functional, tool-oriented tasks, paving the way for more reliable Arabic-native agents.

\section{Related Work}

Since the creation of transformer based neural network architecture~\cite{vaswani2017attention}, the evaluation of LLM tool-calling abilities has become an active area of research, albeit one almost exclusively focused on English. 
Comprehensive efforts like ToolLLM \cite{qin2023toolllm} and ToolBench \cite{guo2024stabletoolbench} provided large-scale datasets and evaluation pipelines, with the latter featuring thousands of real-world APIs. 
Other benchmarks like API-Bank \cite{li2023api} assessed a model's ability to handle complex, multi-step API calls. Furthermore, the \emph{Berkeley Function Calling Leaderboard}~\cite{patil2025bfcl}, stemming from the Gorilla project, has become a widely recognized standard for measuring the accuracy, latency, and instruction-following fidelity of function-calling. 

In parallel, the Arabic NLP community has made significant strides in LLM evaluation~\cite{al2025landscape,wood2024conflict}.
Benchmarks such as ALUE~\cite{seelawi2021alue} and the more recent ARB (Arabic Representation Benchmark)~\cite{ghaboura2025arb} provide extensive testbeds for models across a variety of traditional tasks.
These evaluations typically cover single-turn and multi-turn conversations, question answering, text generation, and embedding quality. 
However, none of these comprehensive Arabic benchmarks include tasks specifically designed to evaluate agentic tool-use.

This focus on non-English tool-use evaluation is a critical and emerging research area. 
While the field has been dominated by English-centric benchmarks, several recent efforts have highlighted the unique challenges of tool-calling in other languages. 
A prime example is FunctionChat-Bench~\cite{lee2024functionchat}, which provides a comprehensive framework for evaluating the generative and tool-use capabilities of models within Korean-language dialogs. 
This benchmark specifically tests a model's ability to handle multi-turn conversations and select the correct Korean-language tools. 
Similarly, a fork of the Berkeley Function Calling Leaderboard has been adapted for Traditional Chinese (zh-tw)~\cite{fc-leaderboard-zhtw}, aiming to address the performance gap seen in multilingual models, which are often undertrained on non-English data. 
Other efforts, such as the work in~\cite{chen2024enhancing}, have explored data and translation strategies to improve multilingual function-calling in languages like Traditional Chinese. 
These works, along with our proposed Arabic benchmark, collectively underscore a vital shift towards ensuring that agentic AI is not only powerful but also linguistically equitable and robust across a global user base.

To our knowledge, work on Arabic tool calling remains nascent. 
The focus has seldom been on benchmarking existing models, but rather on creating data and finetuning models to imbue them with this capability.
For instance, recent work has explored data strategies and instruction tuning specifically to enable tool calling for Arabic LLMs \cite{ersoy2025tool}. 
While this demonstrates the demand for such a feature, it also highlights the lack of a standardized benchmark to measure the \textit{out-of-the-box} tool-calling performance of foundational models when prompted in Arabic. 
Our work is therefore distinct, providing the first public benchmark to measure these agentic workflows directly.

\section{Method}
\label{sec:method}

Our methodology is built upon the Berkeley Function Calling Leaderboard (BFCL)~\cite{patil2023gorilla}, a widely adopted and comprehensive benchmark for evaluating the tool-use capabilities of Large Language Models. 
We extend this benchmark to the standard Arabic language to create a robust framework for multilingual function calling evaluation.

\subsection{Dataset Translation and Preparation}

To adapt the benchmark for Arabic, we developed a multilingual parallel corpus. 
This involved the automated translation of the core components of the BFCL dataset, including user queries, function definitions (with descriptions and parameter names), and system prompts. 
This process was supervised with a minimum human-in-the-loop for quality assurance and to correct any nuanced linguistic errors, ensuring high fidelity to the original dataset's intent.

\subsection{Models Under Evaluation}

We focused our evaluation on a selection of prominent, state-of-the-art open-source models that are widely used and have demonstrated strong performance on general benchmarks. The models selected for this study are:

\subsubsection{GPT-OSS-20b}
This model~\cite{agarwal2025gpt} is a thinking model that is supposed to handle function calling and multi-lingual conversations. 
It is a mixture of experts with 3 billion active parameters.
The exact model weights we are referring to can be found on HuggingFace\footnote{https://huggingface.co/openai/gpt-oss-20b}.

\subsubsection{Llama-3.3-70b}
It was published in~\cite{grattafiori2024llama} as a full Dense non thinking model. 
It is supposed to be multi-lingual yet it does not officially support Arabic, nevertheless our internal quality assessment tests revealed that the model can be used for some Arabic use cases.
The exact model weights we are referring to can be found on HuggingFace\footnote{https://huggingface.co/casperhansen/llama-3.3-70b-instruct-awq}.

\subsubsection{Qwen3}
This model~\cite{yang2025qwen3} exists in many different sizes, we have used two of the most popular sizes for local deployment: \qwenlarge and \qwensmall non-thinking versions.
These models are supposed to be good in function calling and multilingual conversations.
In addition to instruct models, we have included the thinking version of \qwensmall called \qwenthink.
The exact model weights we are referring to can be found on HuggingFace\footnote{https://huggingface.co/collections/Qwen/qwen3}.

\subsection{Experimental Design}
To comprehensively assess multilingual Function Calling (FC) performance, we designed a multi-faceted experimental matrix, detailed in Table~\ref{tab:exp_matrix}. 
This matrix consists of four primary axes, each with two variations, resulting in a total of $2^4 = 16$ unique evaluation combinations for each model and task. 
This design allows us to isolate the impact of language in different parts of the prompt and tool definition.

Here are the paragraph explanations for each of the experimental variations.

\subsubsection{System Prompt Language}

This experiment axis tests the model's sensitivity to the language of its foundational instructions. 
The system prompt sets the model's persona, its core task (to act as an agent), and the rules for tool-use. 
By switching this prompt between \emph{English} and \emph{Arabic}, we can measure if the model's core instruction-following capability and reasoning process are influenced by the language of its primary directive. 
This helps isolate whether a model's \emph{agentic} behavior is language-dependent, even before it processes a user's query.

\subsubsection{Invocation Style}

This axis evaluates the two primary methods for eliciting a tool call from an LLM. \emph{Native Function Calling} utilizes the model's built-in, fine-tuned API feature (e.g., \emph{tool-use}), which is highly optimized to generate structured JSON output. 
In contrast, \emph{Prompt} based invocation tests the model's raw in-context learning. 
In this mode, we instruct the model via the system prompt to generate a JSON object representing the function call directly within its text response, bypassing the specialized API feature. 
This comparison allows us to determine if the model's (likely English-trained) native fine-tuning is more or less robust than its general reasoning when faced with multilingual inputs.

\subsubsection{User Query Language}

This is the central variable of our study, representing the user's direct input.
We test every user query in both its original \emph{English} form and our translated \emph{Arabic} version.
This experimental axis is designed to directly measure the model's ability to understand a user's intent in Arabic and accurately map that intent to the correct tool and its parameters.
Any performance degradation observed when switching from English to Arabic queries highlights the core challenge of multilingual tool-use.

\subsubsection{Function Description Language}

This variable modifies the language of the tool definitions themselves, which are provided to the model as context. 
We test whether providing the function names, descriptions, and parameter details in \emph{English} versus \emph{Arabic} affects the model's selection accuracy. 
This experiment is crucial for determining if performance improves when the query and tool languages match (e.g., Arabic-to-Arabic) or if models perform better by default with English tool definitions, even when handling an Arabic query, by "translating" the query to an internal English representation before tool selection.

\begin{table}[htbp]
\caption{Multilingual Evaluation Matrix}
\begin{center}
\begin{tabular}{|c|c|c|}
\hline
\textbf{Experimental Axis}&\multicolumn{2}{|c|}{\textbf{Experimental Variations}} \\
\cline{2-3} 
\hline
System Prompt Language & English & Arabic \\
\hline
Invocation Style & Native FC & Prompt \\
\hline
User Query Language & English & Arabic \\
\hline
Function Description & English & Arabic \\
\hline
\end{tabular}
\label{tab:exp_matrix}
\end{center}
\end{table}

\subsection{Task Categories}\label{cat-sec}

Our evaluation incorporates the original task categorizations from BFCL to ensure a comprehensive assessment across different function-calling scenarios. 
The categories we utilized are described below:

\subsubsection{Irrelevance Detection}
The model is presented with a user query for which no provided tool is relevant. The model must correctly abstain from calling any function.

\subsubsection{Relevance Detection}
The model is given a list of functions and must determine the single relevant function to call based on the user query.

\subsubsection{Live Simple}
A single function call is required for a given user prompt.

\subsubsection{Live Multiple} 
The prompt requires the model to select the correct function to call from multiple options.

\subsubsection{Live Parallel}
The model must issue multiple function calls in a single turn that can be executed in parallel.

\subsubsection{Live Parallel Multiple}
A combination where the model must make multiple, parallelizable function calls in a single turn.

\subsubsection{Multi Turn Base} 
A baseline multi-turn conversation where the model must maintain context to make a correct function call.

\subsubsection{Multi Turn Long Context} 
A multi-turn scenario with a long conversation history, testing the model's ability to retrieve relevant information from the context.

\subsubsection{Multi Turn Miss Func} 
A multi-turn scenario where a function required by the user is not available, and the model must handle this gracefully.

\subsubsection{Multi Turn Miss Param}
A multi-turn conversation where a necessary parameter for a function call is missing, and the model must ask a clarifying question.

\subsubsection{Non-Live Simple} 
This category corresponds to the \textit{Live Simple} but uses non–user-contributed samples.

\subsubsection{Non-Live Multiple}
This category corresponds to the \textit{Live Multiple} but uses non–user-contributed samples.

\subsubsection{Non-Live Parallel} 
This category corresponds to the \textit{Live Parallel} but uses non–user-contributed samples.

\subsubsection{Non-Live Multiple Parallel} 
This category corresponds to the \textit{Live Multiple Parallel} but uses non–user-contributed samples.

\subsection{Accuracy Computation}

For scoring, we adhere to the original methodology proposed by the BFCL authors \cite{patil2023gorilla}. 
Accuracy is determined by comparing the Abstract Syntax Tree (AST) of the model's generated function call against the ground-truth AST. 
This method ensures a rigorous evaluation of not only the correct function name but also the precise structure and values of the parameters provided.

\subsection{Experimental Hardware Setup}

All experiments were conducted on the Google Cloud Platform (GCP) using a managed Kubernetes (K8s) cluster. 
This environment allowed us to robustly deploy and scale our evaluation workloads.

For model inference, we utilized NVIDIA L4 Tensor Core GPUs, each equipped with 24GB of GDDR6 memory. 
The models were served using vLLM, a high-throughput and memory-efficient serving engine. 
vLLM's implementation of PagedAttention and continuous batching was critical for achieving the performance required to run our extensive set of 16 experimental combinations across all datasets.

Resource allocation was tailored to the size of each model. The smaller models, specifically \gptoss and \qwensmall, were each deployed to a single L4 GPU. 
The larger models, \llama and \qwenlarge, required a multi-GPU configuration. 
For these models, we leveraged vLLM's tensor parallelism capabilities to serve each model across four L4 GPUs. 
Kubernetes was used to manage these deployments and scale the number of replicas as needed to complete the evaluation.

\section{Results}

This section presents the comprehensive results of our large-scale multilingual evaluation. 
To fully assess the impact of language on tool-calling capabilities, we executed a total of 784 distinct experiments. 
This extensive testbed is the product of evaluating four state-of-the-art open-source models across a matrix of linguistic and technical variations.

The models evaluated were the three Qwen models, in addition to \gptoss, and \llama. 
Each model was tested against 14 distinct function-calling categories from the BFCL dataset, covering scenarios such as those mentioned in Section~\ref{cat-sec}.

Our experimental matrix was composed of the three main \emph{Language Variations:} We created 8 unique linguistic combinations based on a $2\times2\times2$ factorial design, varying the language (English or Arabic) of the System Prompt, User Query and Function Description.

To achieve the total of 784 experiments, the models were tested as follows: The \qwenlarge and \qwensmall models were evaluated against all 16 possible combinations (8 language variations $\times$ 2 invocation styles). 
The \gptoss, \qwenthink and \llama models were evaluated against the 8 language variations using their native function-calling capabilities. 
This rigorous, multi-axis evaluation allows us to precisely isolate performance gaps and analyze the specific impact of language on different components of the agentic workflow.

To go beyond simple averages, we conducted a pairwise comparison of all classifiers and experimental variations (e.g., English vs. Arabic user queries) using a direct head-to-head analysis across all experiments.
This involved generating pairwise accuracy plots to visually summarize the aggregate number of \emph{wins}, \emph{losses}, and \emph{ties}.
Furthermore, to validate that the observed performance differences were not merely due to random chance, we supplemented this visual analysis with a formal \emph{statistical test} to compute the \emph{p-value}, thereby establishing the statistical significance of our results using Wilcoxon's signed rank test~\cite{wilcoxon1992individual}.

\subsection{Overall Model Performance by Average Rank}

To synthesize performance across our 672 experiments, we calculated the average rank of each model. For any given experiment (a unique combination of task, language, and invocation style), the models were ranked from 1 (best) to 5 (worst). The \textit{ Prompt} based invocation style was not supported by all models; therefore, the reported average rankings correspond to the \textit{ Native Function Calling} mode. These ranks were then averaged per model.

Table~\ref{tab:avg_rank_overall} shows the overall average rank for each model across all experiments. 
To provide a clearer insight into how language bias affects performance, we break down this ranking based on the language of the user query.
In all tables, a lower score indicates a more consistently high-performing model.

\begin{table}[htbp]
\caption{Average Model Rank}
\begin{center}
\begin{tabularx}{\linewidth}{|l|>{\centering\arraybackslash}X|>{\centering\arraybackslash}X|>{\centering\arraybackslash}X|}
\hline
\textbf{Model} & \textbf{\textit{English Prompts Rank}} & \textbf{\textit{Arabic Prompts Rank}} & \textbf{\textit{Overall Rank}} \\
\hline
\qwenthink & 1.62 & 1.29 & \textbf{1.46} \\
\hline
\qwenlarge & 2.99 & 3.12 & \textbf{3.05} \\
\hline
\llama & 2.78 & 3.35 & \textbf{3.06} \\
\hline
\qwensmall & 3.59 & 3.47 & \textbf{3.54} \\
\hline
\gptoss & 4.02 & 3.77 & \textbf{3.89} \\
\hline
\end{tabularx}
\label{tab:avg_rank_overall}
\end{center}
\end{table}

The results presented in Table~\ref{tab:avg_rank_overall} reveal a clear performance hierarchy among the evaluated models and highlight notable differences between English and Arabic prompt handling.

As shown in the \textit{English Prompts Rank} column, \qwenthink stands out as the top performer on English prompts, achieving an average rank of \textbf{1.62}, well ahead of the other models. 
\llama and \qwenlarge follow in the next tier, with average ranks of \textbf{2.78} and \textbf{2.99}, respectively. 
In contrast, \qwensmall and \gptoss trail behind, with ranks of \textbf{3.59} and \textbf{4.02}, indicating weaker performance on English-language tasks.

When focusing on Arabic prompts (\textit{Arabic Prompts Rank} column), \qwenthink again leads with an average rank of \textbf{1.29}, which is even higher than in English. 
However, this improvement likely reflects a \textbf{relative decline in the performance of other models} — particularly \llama and \qwenlarge, whose average ranks worsen to \textbf{3.35} and \textbf{3.12}, respectively—rather than clear evidence that \qwenthink performs better on Arabic itself. \qwensmall and \gptoss show minimal changes between languages, ranking \textbf{3.47} and \textbf{3.77}, suggesting relatively stable but lower performance across both prompt sets.

\textit{Overall Prompts Rank} column clearly shows \qwenthink achieves the best combined performance across both languages with an average rank of \textbf{1.46}, outperforming all competitors by a substantial margin. The mid-tier models, \qwenlarge and \llama, cluster closely around \textbf{3.05}–\textbf{3.06}, while \qwensmall and \gptoss occupy the lower end of the ranking spectrum.

These findings suggest that \qwenthink delivers consistently superior results across both English and Arabic, whereas other models—particularly \qwenlarge and \llama — exhibit a stronger bias toward English prompts.
Meanwhile, although \qwensmall and \gptoss display more uniform average rank performance, it does not indicate a less pronounced language bias, but rather a very low rank for both languages.

It is also worth noting that the \gptoss model’s relatively low standing is likely influenced by its inability to perform parallel function calls, a constraint inherited from its fine-tuning setup.

Further post-hoc statistical analysis is needed to determine whether these observed ranking differences are statistically significant or simply reflect sample variability.

\subsection{Pairwise Model Comparisons}

\begin{figure}[htbp]
\centerline{\includegraphics[width=0.9\columnwidth]{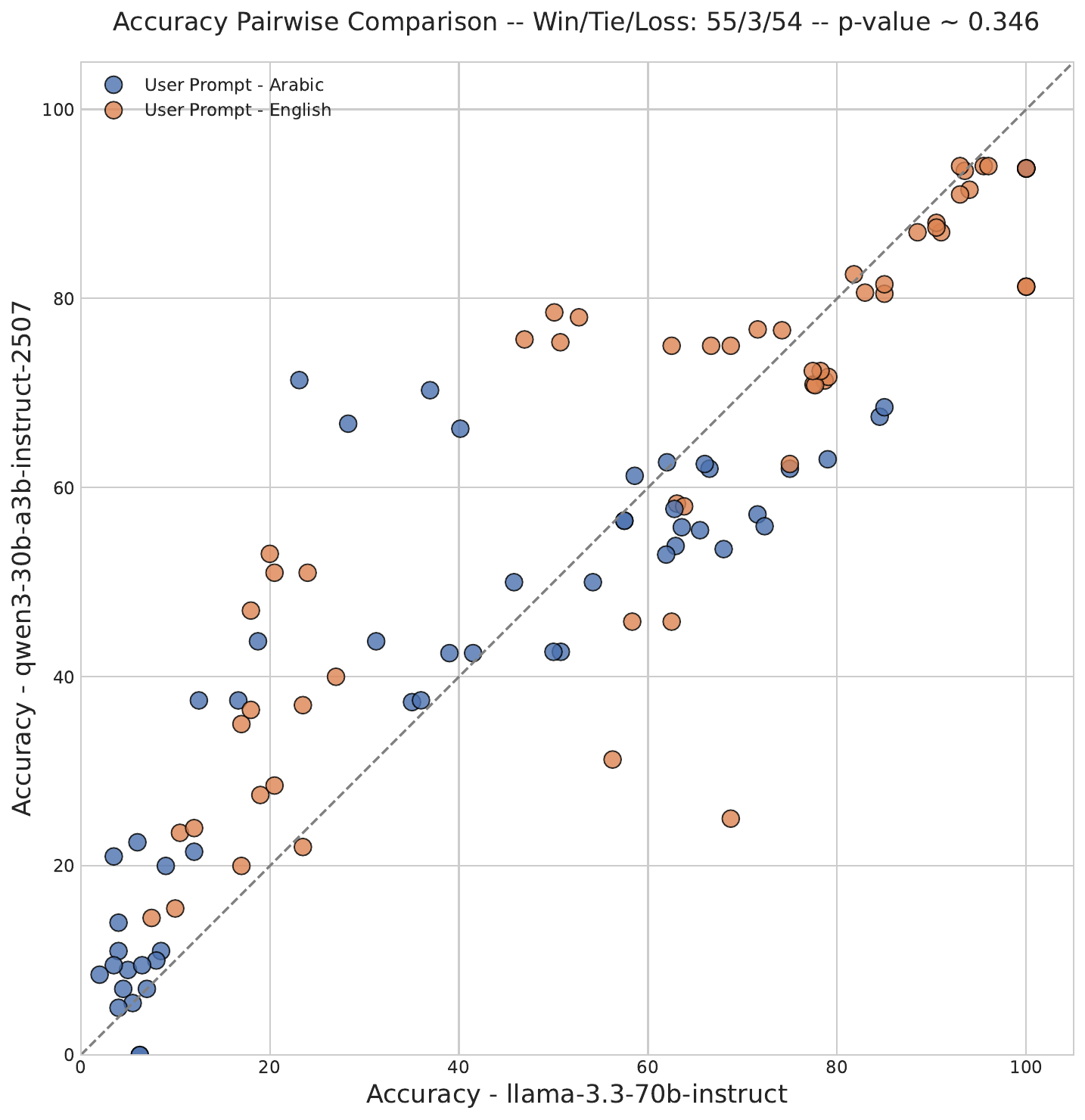}}
\caption{Accuracy comparison between \llama (x-axis) and \qwenlarge (y-axis).}
\label{fig:llama-vs-qwen30}
\end{figure}

\begin{figure}[htbp]
\centerline{\includegraphics[width=0.9\columnwidth]{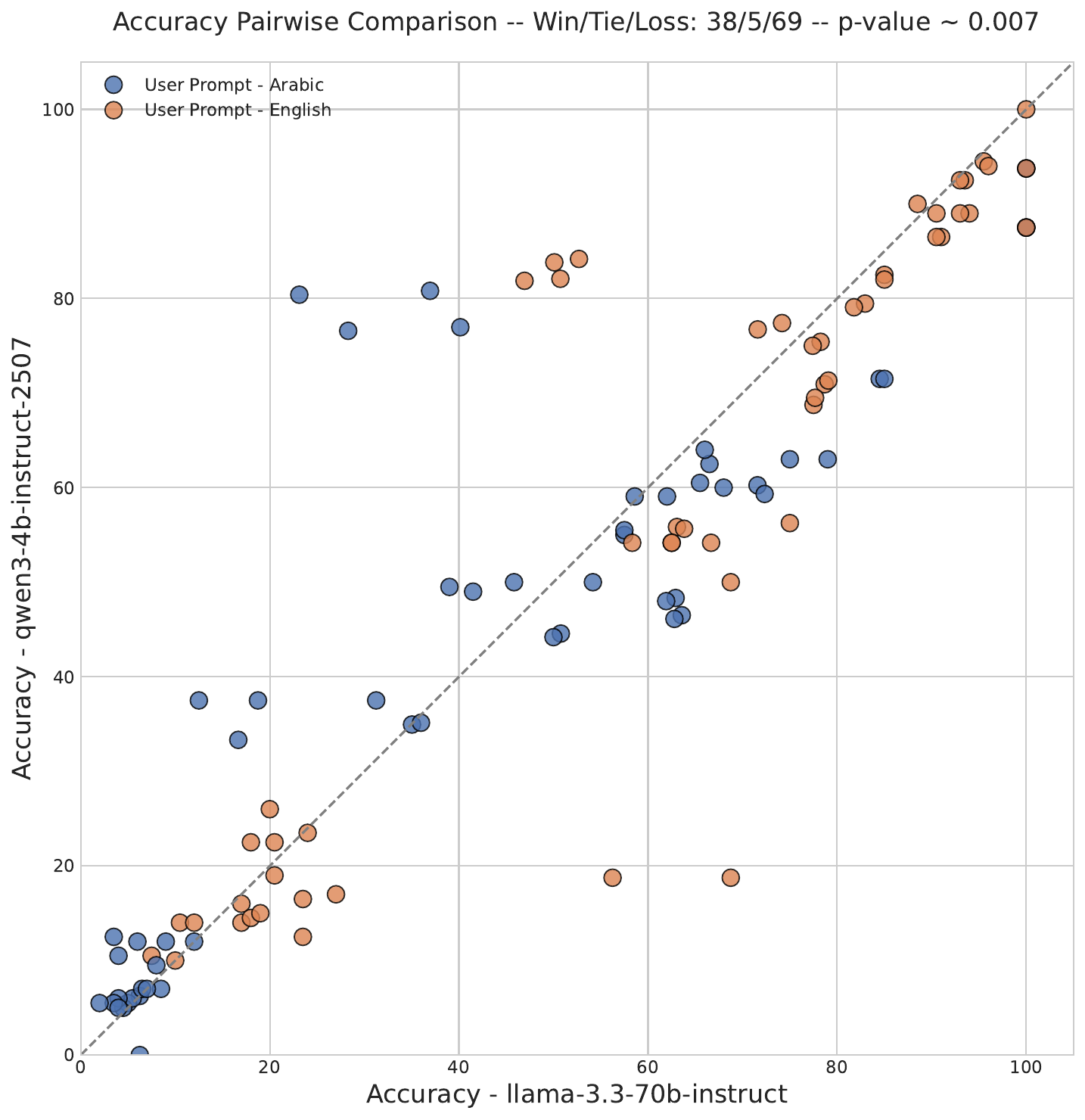}}
\caption{Accuracy comparison between \llama (x-axis) and \qwensmall (y-axis).}
\label{fig:llama-vs-qwen4}
\end{figure}

\begin{figure}[htbp]
\centerline{\includegraphics[width=0.9\columnwidth]{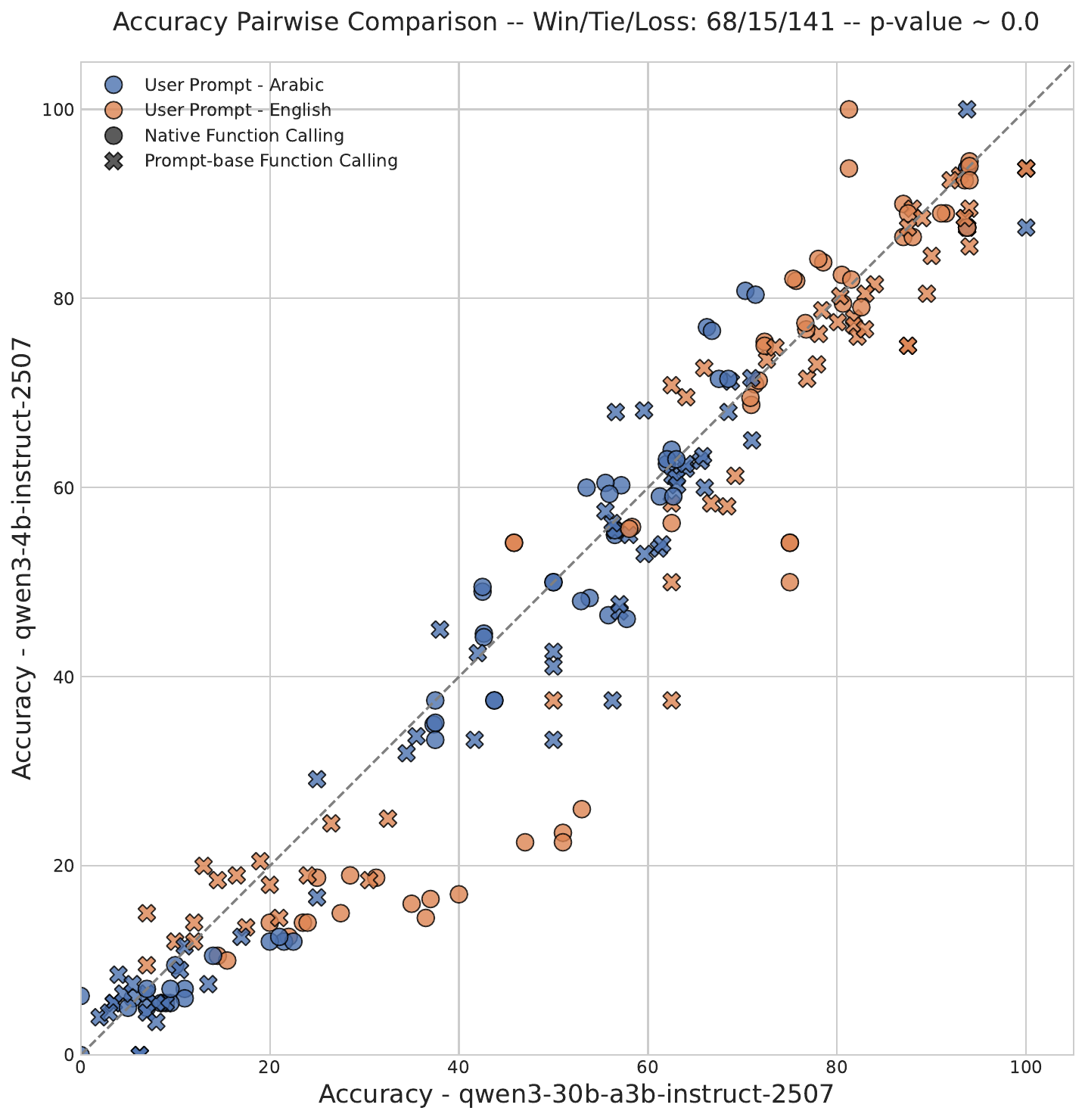}}
\caption{Accuracy comparison between \qwenlarge (x-axis) and \qwensmall (y-axis). Note the 'x' markers for prompt-based invocation.}
\label{fig:qwen30-vs-qwen4}
\end{figure}

\begin{figure}[htbp]
\centerline{\includegraphics[width=0.9\columnwidth]{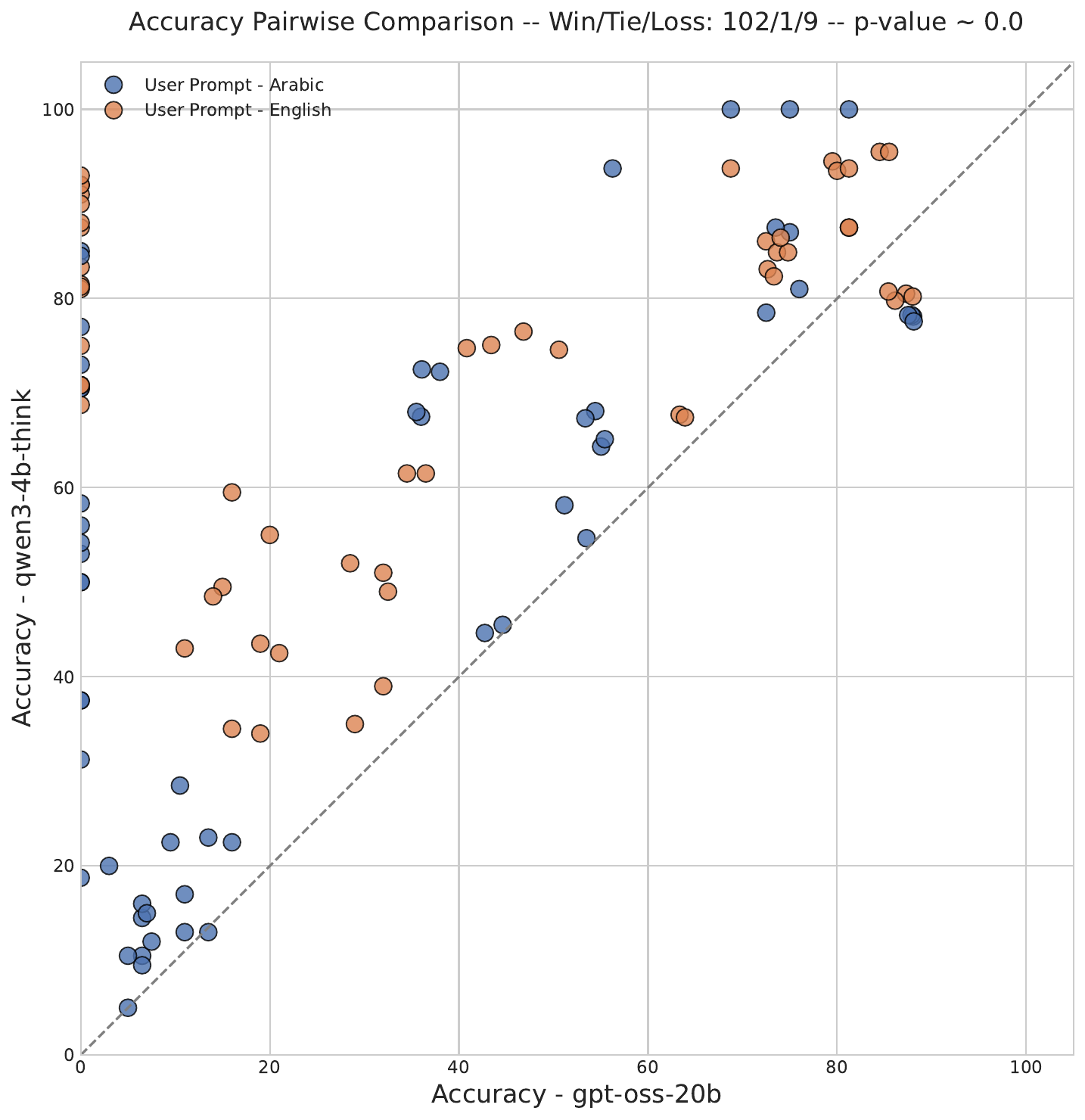}}
\caption{Accuracy comparison between \gptoss (x-axis) and \qwensmall (y-axis).}
\label{fig:gpt-vs-qwen4}
\end{figure}

\begin{figure}[htbp]
\centerline{\includegraphics[width=0.9\columnwidth]{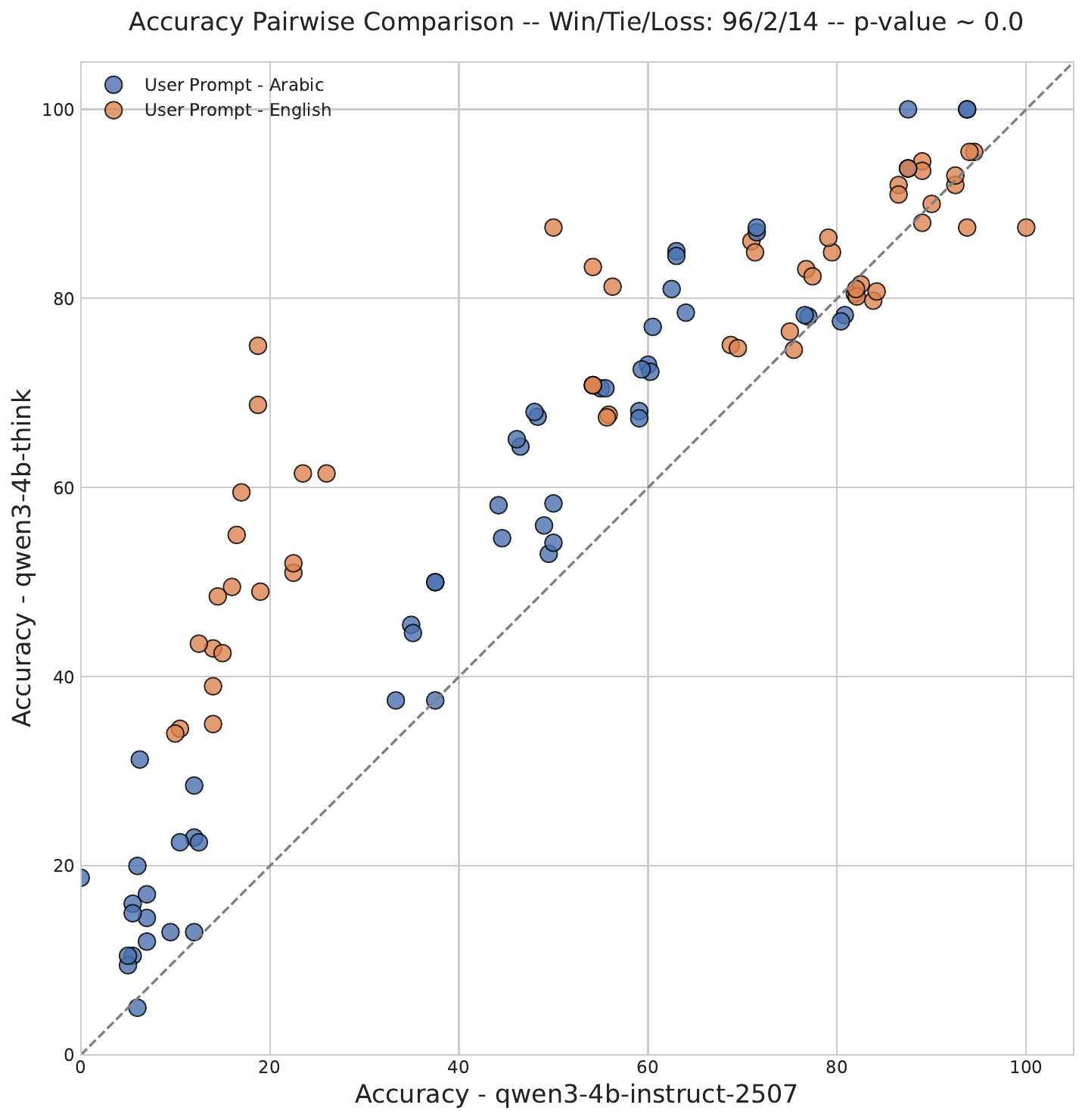}}
\caption{Accuracy comparison between \qwensmall (x-axis) and \qwenthink (y-axis).}
\label{fig:qwen3-4b-instruct-vs-qwen3-4b-thinking}
\end{figure}

To visualize the direct head-to-head performance of the models, we generated pairwise accuracy scatter plots. 
In Figures \ref{fig:llama-vs-qwen30} through \ref{fig:qwen3-4b-instruct-vs-qwen3-4b-thinking}, each point represents the accuracy of the two models on a single experimental setup (a specific task and language/style variation).

Points above the diagonal line indicate a "win" for the model on the y-axis, while points below the line indicate a "win" for the model on the x-axis. The Win/Tie/Loss (W/T/L) counts in the titles follow a \textit{Y-axis wins / Ties / X-axis wins} convention. 
A p-value is included to determine if the difference in the win/loss distribution is statistically significant, with a threshold of $\le 0.05$ for significance level.

Overall the pairwise plots confirm the hierarchy established in our average rank analysis (Table~\ref{tab:avg_rank_overall}).

\textbf{\llama vs. \qwenlarge (Fig.~\ref{fig:llama-vs-qwen30}):}
We first compare the two largest models. 
The results show a statistical dead heat. 
The points are scattered very evenly around the diagonal line, with \qwenlarge (y-axis) winning 55 experiments and \llama (x-axis) winning 54. 
The high p-value of $\sim 0.346$ confirms that there is \textbf{no statistically significant difference} between them.
Furthermore, we can clearly see that when it comes to Arabic user prompts, both models exihibt a huge drop in accuracy where go from a maximum of 100\% (on a couple of English user prompts runs for Llama) to a maximum of around 84\% (for the best Arabic user prompt experiment for Llama).
Same observation for \qwenlarge showing that both models are suffering from a significant drop in accuracy when user prompts are in Arabic.
In summary, the choice between these two models will not be based on accuracy but will rather be based on speed and memory usage, therefore the preferred model would be \qwenlarge since it is a mixture of experts model. 

\textbf{\llama vs. \qwensmall (Fig.~\ref{fig:llama-vs-qwen4}):}
This plot compares the largest model to the smallest.
\llama (x-axis) wins decisively, securing 69 wins versus only 34 for \qwensmall (y-axis). 
The low p-value for the Wilcoxon signed rank test of $\sim 0.005$ indicates this performance gap is statistically significant.
Similarly to the previous comparison, an apparent drop in terms of accuracy is observed when it comes to experiments with Arabic user prompts: \qwensmall goes from a maximum of 100\% accuracy (for English user prompt) to a maximum of 82\% (for Arabic user prompt). 
Finally, in terms of accuracy, we can clearly state that the choice of \llama as our tool calling model triumphs its rival. 
Nevertheless, we are required to compare both \qwensmall and \qwenlarge since the initial comparison in Fig.~\ref{fig:llama-vs-qwen30} convinced us to choose \qwenlarge over \llama.

\textbf{\qwenlarge vs. \qwensmall (Fig.~\ref{fig:qwen30-vs-qwen4}):}
We observe the same "bigger is better" trend within the Qwen family.
The larger \qwenlarge (x-axis) model significantly outperforms its smaller 4B sibling (y-axis), with 141 wins to 68. 
The p-value of $\sim 0.0$ underscores this. 
Notably, many of the prompt-based invocation experiments (marked with an 'x') fall far below the diagonal, suggesting the 30B model is particularly more adept at prompt-based tool-use than the 4B model.
However the focus of this benchmark is the behavior of these models when it comes to Arabic language (specifically Arabic user prompts). 
In this case, we can clearly observe that the winning rate of \qwenlarge is heavily impacted by the English language (yellow dots in Fig.~\ref{fig:qwen30-vs-qwen4}). 
This does not mean that \qwensmall is better in Arabic than \qwenlarge but rather that both are equally bad and perhaps \qwenlarge is more finetuned on English tool calling datasets and is able to capture this type of knowledge compared to its smaller counterpart \qwensmall. 
In general in a multilingual setup, the impact that we observed from our own internal manual testing, that \qwensmall is not significantly different when it comes to tool calling in practice, and given the cost of deployment is 4 times higher, we recommend GenAI practitioners to start testing with \qwensmall before going with larger options such as \qwenlarge. 

\textbf{\gptoss vs. \qwenthink (Fig.~\ref{fig:gpt-vs-qwen4}):}
We continue by comparing the two thinking (reasoning) models. 
Here, the \qwenthink (y-axis) achieves a clear and statistically significant victory over \gptoss (x-axis), with 102 wins to 9. 
This result solidifies the overall performance ranking, confirming that \qwenthink is a stronger performer than its reasoning counterpart \gptoss for function calling in both languages.
However we should note that the low performance for \gptoss stems from the fact that this model completely fails in all parallel function calling datasets. 
The limitation is observed by other practitioners that tried leveraging \gptoss for parallel function calling but unfortunately it seems like OpenAI did not finetune this model to perform multiple tool calling in parallel thus explaining the vertical dots (experiments) for $x=0$ where \gptoss accuracy is zero.
Nevertheless, even when we ignore the $x=0$ axis, we can still see that \qwenthink is more accurate than \gptoss on average.

\textbf{\qwenthink vs \qwensmall (Fig.~\ref{fig:qwen3-4b-instruct-vs-qwen3-4b-thinking})}

To assess the effect of leveraging a reasoning model vs an instruct one, we compared the standard \qwensmall model (x-axis) against its "thinking" variant (\qwenthink, y-axis). 

The plot clearly demonstrates that the reasoning version of \qwensmall presents a substantial performance improvement. 
The majority of points fall above the diagonal line, indicating superior accuracy for the think variant. 
The Win/Tie/Loss count is decisively in favor of \qwenthink, with 96 wins compared to only 14 for the instruct model, and 2 ties.
This difference is highly statistically significant (p-value $\sim 0.0$).

Observing the language distribution (colors), the improvement is evident for both English (orange) and Arabic (blue) prompts. 
While the instruct model struggles significantly, particularly at lower accuracy levels on Arabic prompts (blue dots clustered near the origin on the x-axis), the reasoning variant consistently achieves higher accuracy on these same tasks (corresponding blue dots higher on the y-axis). 
This suggests that incorporating thinking process into the post-training regime of LLMs, significantly enhances the model's ability to handle function-calling tasks, particularly boosting performance on the more challenging Arabic user queries.

\subsection{Analysis of Experimental Variations}
\label{sec:pairwise_variations}

To isolate the effect of each of the four experimental axes, we conducted a head-to-head analysis by plotting the accuracy of one variation (e.g., English) against the other (e.g., Arabic). 
In the following plots, each point represents the accuracy for a specific task and model combination. 
Points below the diagonal dashed line represent a "win" for the variable on the x-axis, while points above represent a "win" for the y-axis. 
The Win/Tie/Loss (W/T/L) counts are reported in the (Y-axis wins / Ties / X-axis wins) format.

\subsubsection{Impact of User Prompt Language}

\begin{figure}[htbp]
\centerline{\includegraphics[width=0.9\columnwidth]{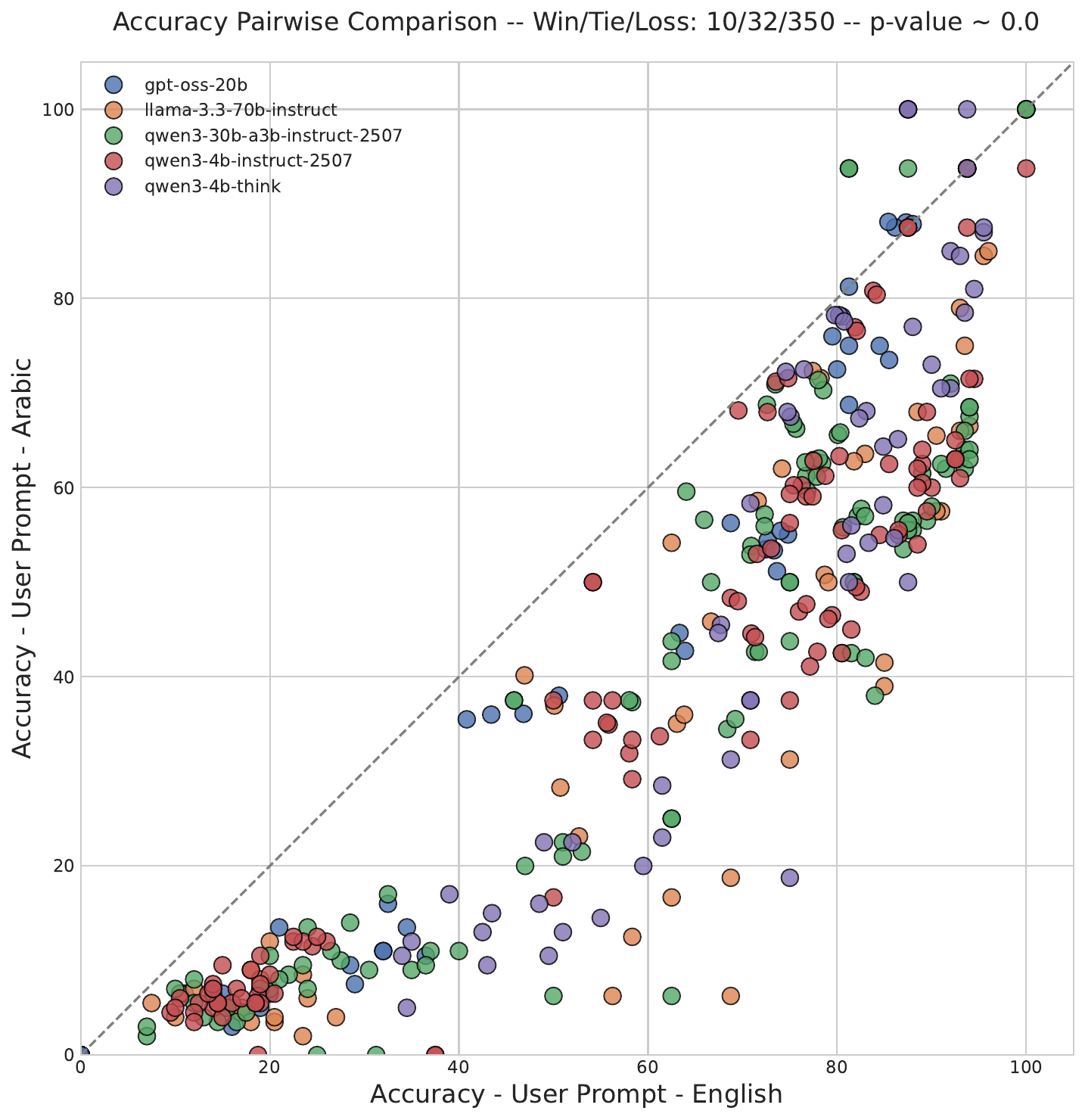}}
\caption{Accuracy comparison for English (x-axis) vs. Arabic (y-axis) user prompts.}
\label{fig:user_prompt_lang}
\end{figure}

Figure~\ref{fig:user_prompt_lang} reveals the most significant finding of our entire study. 
The comparison between English (x-axis) and Arabic (y-axis) user queries shows a dramatic and near-total performance collapse across all tested models. 
The plot is overwhelmingly skewed below the diagonal, with English-based prompts winning in 350 test cases, versus a mere 10 wins for Arabic-based prompts. 
The p-value of $\sim 0.0$ confirms that this degradation is highly statistically significant.

This visualization starkly illustrates the 5-10\% average performance drop we observed. 
In many instances, the drop is far more severe; for example, tasks where models achieve 80-90\% accuracy with English prompts (x-axis) often see their performance plummet to 40-60\% when given the equivalent Arabic prompt (y-axis).
This demonstrates that all models in our evaluation, regardless of their size or family, struggle severely to understand and execute tool-use tasks when the user's intent is expressed in Arabic.

\subsubsection{Impact of Function Description Language}

\begin{figure}[htbp]
\centerline{\includegraphics[width=0.9\columnwidth]{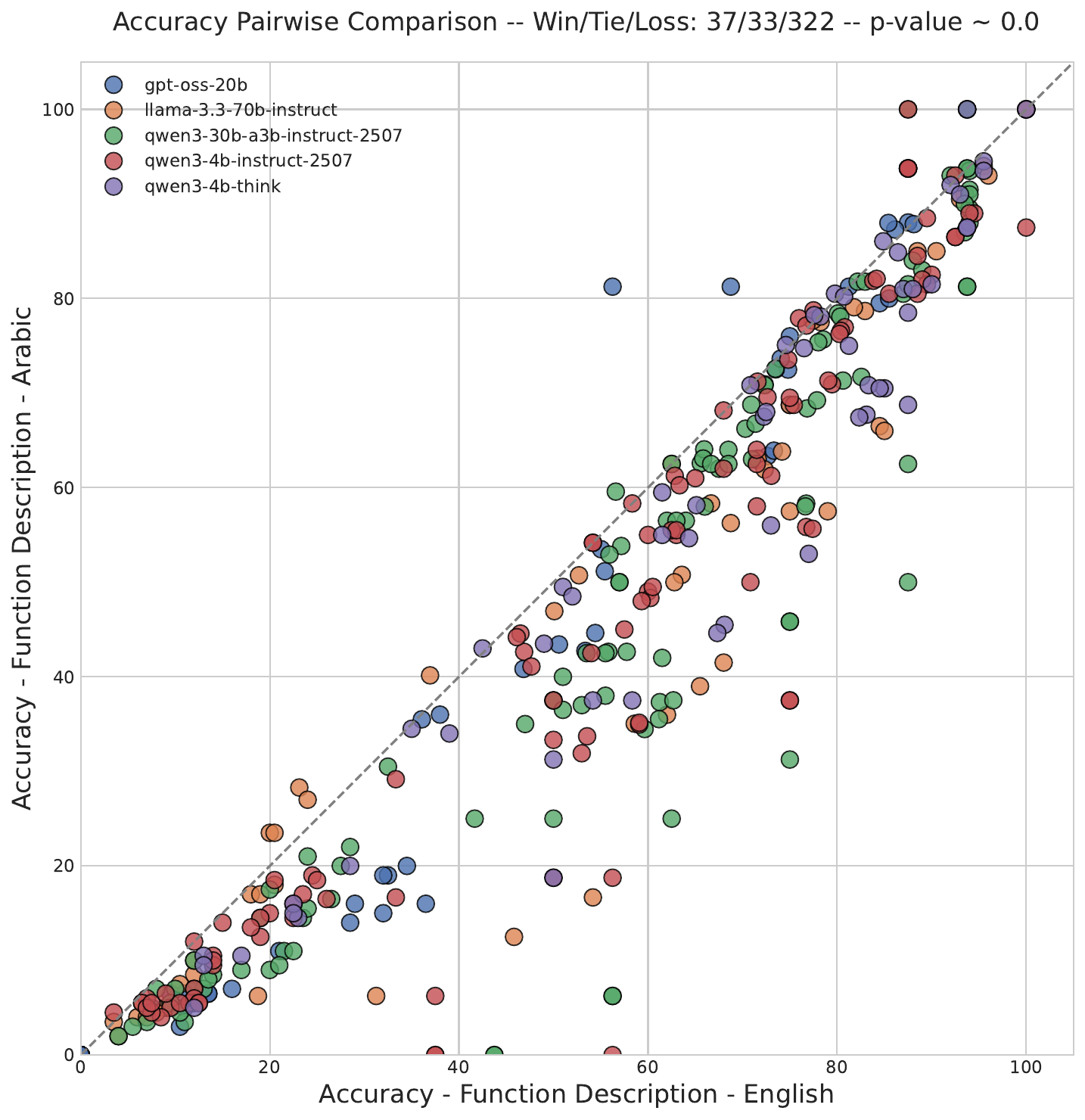}}
\caption{Accuracy comparison for English (x-axis) vs. Arabic (y-axis) function descriptions.}
\label{fig:func_desc_lang}
\end{figure}

A similar, albeit slightly less severe, trend is observed when comparing the language of the tool definitions themselves.
As shown in Figure~\ref{fig:func_desc_lang}, providing the function and parameter descriptions in English (x-axis) is vastly superior to providing them in Arabic (y-axis).

English-based descriptions led to 322 wins, compared to only 37 for Arabic, with a p-value of $\sim 0.0$ which is also statistically significant. 
This finding suggests that the models' tool-use capabilities are not only biased toward English user queries but also heavily rely on English-language context for the tools themselves.
This implies that even if a model correctly understands an Arabic user query, it may fail to select the correct tool if the tool's own definition is also in Arabic.
In summary, attempting to "localize" the entire ecosystem by translating tool definitions to Arabic resulted in a significant, measurable performance loss.

\subsubsection{Impact of System Prompt Language}

\begin{figure}[htbp]
\centerline{\includegraphics[width=0.9\columnwidth]{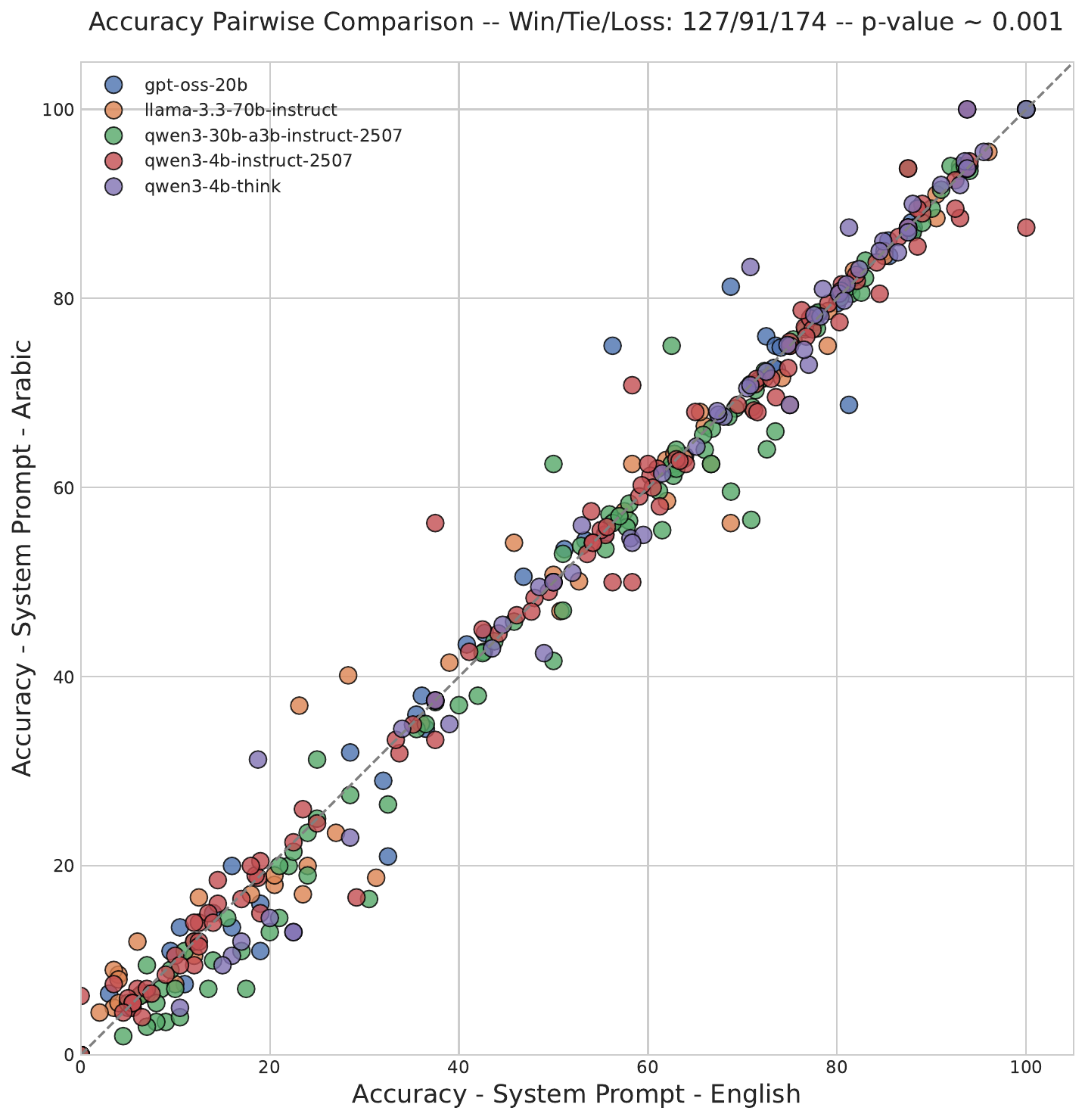}}
\caption{Accuracy comparison for English (x-axis) vs. Arabic (y-axis) system prompts.}
\label{fig:sys_prompt_lang}
\end{figure}

The impact of the system prompt's language, which sets the model's core "agentic" instructions, is shown in Figure~\ref{fig:sys_prompt_lang}. 
Here, the results are far more balanced, with data points clustered much closer to the diagonal.

The p-value of $\sim 0.005$ confirms this trend is not due to random chance. 
This indicates that the language of the model's foundational instructions has a measurable negative effect on its agentic performance, though it is far less severe than the impact of the user query or the function descriptions, as reflected by the corresponding Win/Tie/Loss results.

\subsubsection{Impact of Invocation Style (Native FC vs. Prompt-based)}

\begin{figure}[htbp]
\centerline{\includegraphics[width=0.9\columnwidth]{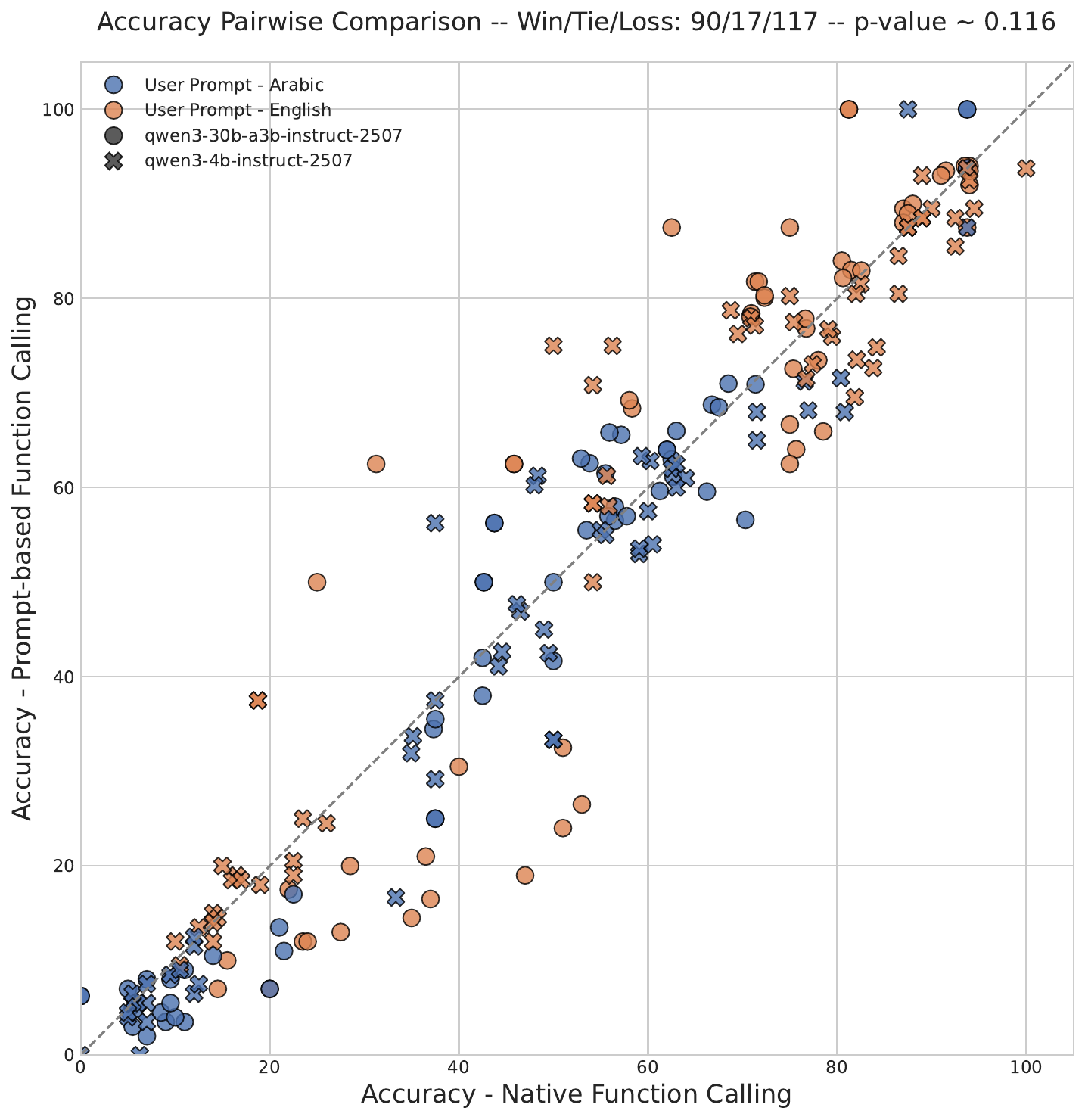}}
\caption{Accuracy comparison for Native Function Calling (x-axis) vs. Prompt-based (y-axis) invocation on Qwen models.}
\label{fig:fc_vs_prompt}
\end{figure}

Finally, we compared the model's native function calling (`fc`, x-axis) against prompt-based tool-use (`prompt`, y-axis). 
This experiment was conducted on the \qwenlarge{} and \qwensmall{} models. 
As seen in Figure~\ref{fig:fc_vs_prompt}, the results are very evenly distributed around the diagonal.
While native function calling secured 117 wins to the prompt-based method's 90, the high p-value of $\sim 0.118$ indicates there is \textbf{no statistically significant difference} between the two methods. 
The plot legend further shows this holds true for both Arabic (`ar`) and English (`en`) user prompts. 
This suggests that, at least for the Qwen models, their general-purpose reasoning (prompt-based) is just as effective as their specialized function-calling fine-tuning for these tasks.
Nevertheless in a production setup, it is highly recommended to utilize the native FC capabilities to avoid model's hallucination and ensure API stability.

\subsection{The Compounding Language Penalty}
\label{sec:compounding_penalty}

\begin{figure}[htbp]
\centerline{\includegraphics[width=0.9\columnwidth]{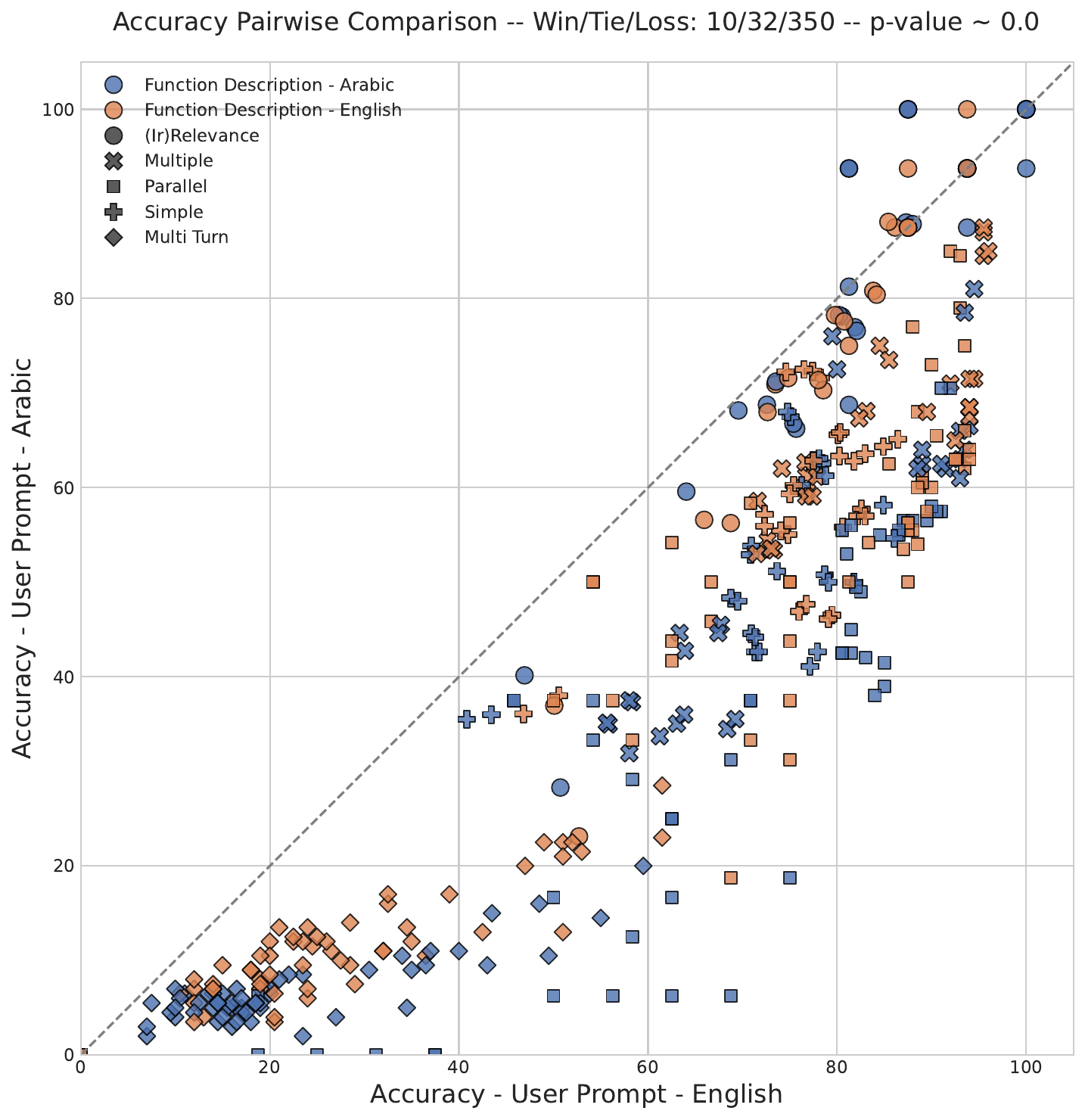}}
\caption{Accuracy comparison for English (x-axis) vs. Arabic (y-axis) user prompts, broken down by function description language (color) and task category (shape).}
\label{fig:user_prompt_dataset}
\end{figure}

While the previous sections established a clear performance hierarchy between models, Figure~\ref{fig:user_prompt_dataset} allows us to dissect the \textit{source} of the performance degradation across our entire experimental suite. 
This plot visualizes the head-to-head comparison of English user prompts (x-axis) versus Arabic user prompts (y-axis). 
The result is a stark confirmation of our primary hypothesis: the models' performance collapses when users interact in Arabic.

The plot is overwhelmingly skewed below the diagonal, with English-based prompts winning in 350 of the scenarios, versus a mere 10 wins for Arabic-based prompts. 
With 32 ties, the p-value of $\sim 0.0$ confirms this is a highly statistically significant finding.

However, this plot reveals a more nuanced and critical insight through its color-coding of the function description language. 
This allows us to isolate two distinct types of multilingual failure. 
First, we observe a \textbf{Query Translation Failure}, represented by the orange points (representing English function description). 
These experiments, where the user query is in Arabic but tool definitions remain in English, are already far below the diagonal, confirming that models struggle to map Arabic intent to English tools. 
Second, we see a \textbf{Compounding Failure}, represented by the blue points (representing Arabic function description). 
In this "fully localized" environment, where both query and tool definitions are in Arabic, a blue point is in nearly every case at the same vertical level or \textit{even lower} than its orange counterpart at the same x-axis position.

This "compounding language penalty" is a key finding. 
It speculates that the models' internal logic for tool-use is fundamentally English-centric. 
They are not effective at matching Arabic intent to Arabic definitions. 
The best (though still poor) performance is achieved when the model can (presumably) translate the Arabic query internally and match it against the English tools it was trained on. 
Forcing the model to \textit{also} parse and understand Arabic tool definitions introduces a second layer of error, resulting in a further performance drop.

\subsection{Task-Specific Fragility}

Figure~\ref{fig:user_prompt_dataset}'s shape-coding, representing the dataset category, reveals where this penalty is most severe. 
Based on a full manual inspection of raw results, we grouped the BFCL tasks into the following meta-categories:

\begin{itemize}
    \item (Ir)Relevance: Containing both Relevance and Irrelvance 
    \item Parallel: Containing all parallel function calling tasks
    \item Multi Turn: Containing all multi-turn BFCL tasks
    \item Simple: Containing all Simple (Live and Non-Live) tasks
    \item Multiple: Containing multiple sequential FC tasks
\end{itemize}

\textbf{Multi-Turn Tasks}, represented by diamonds, are clustered at the bottom-left of the plot.
This indicates they are exceptionally difficult for all models, with low accuracy in both English (often $< 40\%$) and even lower accuracy in Arabic (often $< 20\%$). 
The added cognitive load of a multi-turn conversation combined with a non-English language makes these tasks almost unusable.

This fragility is also pronounced in \textbf{Complex Agentic Tasks}. 
The `Parallel` (squares) and `Multiple` (crosses) function call scenarios, which are the cornerstone of complex agents, show the most dramatic practical performance drop. 
These tasks have high accuracy in English (x-axis, 70-95\%) but suffer a severe and consistent drop when switched to Arabic (y-axis, 50-75\%). 
In contrast, \textbf{Relevance \& Simple Tasks}, marked by circles and pluses, appear to be the most robust.
While still showing a clear drop, they have the most points clustered relatively close to the diagonal. 
This suggests that identifying \textit{if} a tool is needed is a simpler, more language-agnostic task than determining \textit{how} to execute complex, parallel, or sequential tool calls.

In summary, the performance hit from Arabic user queries is universal, but it is significantly compounded when tool definitions are also localized. 
This fragility is most pronounced in the complex, multi-step agentic scenarios that are most critical for real-world applications.

\section{Conclusion, Limitation \& Future Work}

This paper introduced the first dedicated benchmark for evaluating the tool-calling and agentic capabilities of Large Language Models (LLMs) in the Arabic language, addressing a critical gap in the predominantly English-centric evaluation landscape. 
By adapting the widely-used Berkeley Function Calling Leaderboard (BFCL) through a meticulous translation process and employing a comprehensive multi-axis experimental design across five prominent open-source models, we systematically investigated the impact of language on various components of the agentic workflow. 
Our results unequivocally demonstrate a significant performance deficit when models interact with users in Arabic. 
We observed a consistent and statistically significant drop in tool-calling accuracy, averaging between 5-10\% and often much higher in specific scenarios, merely by switching the user query language from English to Arabic. 
This degradation occurred regardless of whether the tool descriptions were provided in English or Arabic, highlighting a deep-seated bias towards English within these models' reasoning processes. 
Furthermore, we identified a "compounding language penalty," where using both Arabic queries and Arabic tool descriptions led to even worse performance than using Arabic queries with English tools, suggesting the models' internal tool-matching logic is fundamentally English-based. 
This fragility was particularly pronounced in complex, multi-step, and parallel tool-use scenarios, which are crucial for real-world agentic applications. 
Our benchmark and findings provide a crucial baseline for the community, underscoring the urgent need for focused efforts to develop and evaluate more linguistically equitable and robust Arabic AI agents. 

Despite the comprehensive evaluation, this study has several limitations. The benchmark relies largely on automated translation (with minimal human review) of the English BFCL dataset, which may not fully capture Arabic nuances, cultural context, or linguistic ambiguities, potentially affecting query and function interpretation. Our model selection includes only five open-source architectures and sizes, limiting generalization to other LLMs, particularly closed-source or Arabic/multilingual fine-tuned models. By building on BFCL, the benchmark inherits its task types and structures, omitting agentic tasks or interaction patterns specific to Arabic contexts. Finally, the Abstract Syntax Tree based evaluation metric ensures syntactic correctness but may not fully capture semantic validity, especially for complex or free-form Arabic parameters.

Addressing the limitations of this study opens several avenues for future research crucial for bridging the observed English-Arabic performance gap in tool-calling. 
A primary direction is the creation of native Arabic tool-calling benchmarks and datasets, developed from scratch to reflect authentic Arabic user intents, linguistic structures, and culturally relevant tasks, rather than relying solely on translation.
Therefore collaborations with local entities such as King Saud University would provide a more genuine measure of Arabic capabilities. 
Secondly, targeted research into multilingual fine-tuning strategies specifically for tool-use is essential. 
This could involve developing cross-lingual alignment techniques for function signatures, creating high-quality parallel instruction-tuning data for tool calling in multiple languages including Arabic, or exploring methods to explicitly teach models to handle code-switching within tool-use dialogues. 
Furthermore, developing more sophisticated, Arabic-aware evaluation metrics beyond exact AST matching is needed. 
Techniques leveraging semantic similarity for parameter values, understanding Arabic morphological variations, or even employing LLM-based evaluators specifically prompted for Arabic functional correctness could provide a more nuanced assessment. 
Expanding the benchmark to include a wider variety of models, especially Arabic-native models and leading closed-source systems, would also provide a more complete picture of the landscape. 
Finally, deeper investigation into the root causes of the performance degradation — whether stemming from tokenization inefficiencies, biases in pre-training data, or suboptimal internal representations for non-English languages — could guide the development of fundamentally more multilingual agentic architectures.

\bibliographystyle{IEEEtran}
\bibliography{ref}

\end{document}